# Boosted Convolutional Neural Networks for Motor Imagery EEG Decoding with Multiwavelet-based Time-Frequency Conditional Granger Causality Analysis

Yang Li, Mengying Lei*, Xianrui Zhang, Weigang Cui, Yuzhu Guo, Ting-Wen Huang, and Hua-Liang Wei*

*Abstract*—Decoding EEG signals of different mental states is a challenging task for brain-computer interfaces (BCIs) due to non-stationarity of perceptual decision processes. This paper presents a novel boosted convolutional neural networks (ConvNets) decoding scheme for motor imagery (MI) EEG signals assisted by the multiwavelet-based time-frequency (TF) causality analysis. Specifically, multiwavelet basis functions are first combined with Geweke spectral measure to obtain high-resolution TF-conditional Granger causality (CGC) representations, where a regularized orthogonal forward regression (ROFR) algorithm is adopted to detect a parsimonious model with good generalization performance. The causality images for network input preserving time, frequency and location information of connectivity are then designed based on the TF-CGC distributions of alpha band multi-channel EEG signals. Further constructed boosted ConvNets by using spatio-temporal convolutions as well as advances in deep learning including cropping and boosting methods, to extract discriminative causality features and classify MI tasks. Our proposed approach outperforms the competition winner algorithm with 12.15% increase in average accuracy and 74.02% decrease in associated inter subject standard deviation for the same binary classification on BCI competition-IV dataset-IIa. Experiment results indicate that the boosted ConvNets with causality images works well in decoding MI-EEG signals and provides a promising framework for developing MI-BCI systems.

## I. INTRODUCTION

THE electroencephalogram (EEG) based brain-computer interface (BCI) is a state-of-the-art technology which establishes a direct communication pathway between human brain and external devices by translating neuronal activities into a series of output commands to accomplish user's intentions [1], and thereby has a wide range of applications from clinic to industry for both patients and normal people [2], such as controlling wheelchair or prosthesis to improve the disabled life quality [3], affecting neural plasticity to facilitate stroke rehabilitation [4], and handling computer games for entertainment of healthy users [5]. Despite the impressive advancements in recent years, EEG-BCI technology is still not able to decode complicated human mental state because of the high complexity of cognitive processing procedure in brain and low signal-to-noise ratio in EEG signals. Hence it is necessary to develop an effective EEG decoding scheme for enhancing usability and interpretability of BCI systems.

Analyzing the EEG signals induced by motor imagery (MI) is one of the most popular but challenging paradigm in BCIs [6]. The key step for MI-BCI implementations is to use machine learning techniques to extract information from EEG recordings of brain activities [7]. Among various types of feature representations for MI-EEG decoding, connectivity patterns of multi-channel signals could generate more discriminating features compared with static single-channel derived features [8] such as the well-known common spatial patterns (CSP) [9, 10], since the dynamic and oscillatory interactions among different regions in the sensorimotor cortex of brain play a fundamental role in accomplishing movement imaginations [11, 12]. Over the latest few years, several approaches have been proposed to analyze connectivity-based MI-BCI systems [13]. For example, Billinger *et al.* [14] suggested a method to extracting single-trial directed transfer functions (DTF) from vector autoregressive (VAR) models of independent components for MI-BCI classification, where the classification results were similar to band power (BP) features. In the work of Rathee *et al.* [15], time-domain partial Granger causality (PGC) is used as the connectivity features in a MI-BCI setting, and it turned out that single-trial effective connectivity distribution can enhance discriminability of mental imagery tasks. In general, the connectivity measures mentioned above can produce useful discriminant features for the classification of brain responses evoked during certain tasks. However, these methods, which assume the stationarity of EEG signals, cannot disclose important dynamic temporal information of connectivity, thus fail to provide robust distinction for nonstationary and complex MI-EEGs, and further result in dissatisfactory classification results with commonly used classification algorithms such as support vector machine (SVM).

Compared to conventional DTF and PGC methods, the time-varying Granger causality (GC) analysis [16], which has proven to be effective for detecting dynamic directed interaction

Manuscript received September 24, 2018. This work was supported by the National Natural Science Foundation of China [61671042, 61403016], and Beijing Natural Science Foundation [4172037]. (Corresponding authors: Mengying Lei; Hua-Liang Wei.)

Yang Li is with the Department of Automation Sciences and Electrical Engineering, Beijing Advanced Innovation Center for Big Data and Brain Computing, Beijing Advanced Innovation Center for Big Date-based Precision Medicine, Beihang University, Beijing, China.

Mengying Lei, Xianrui Zhang, Weigang Cui, and Yuzhu Guo are with the Department of Automation Sciences and Electrical Engineering, Beihang University, Beijing, China (e-mail: lmylei@buaa.edu.cn).

Ting-Wen Huang is with the Department of Mathematics, Texas A&M University at Qatar, Doha 23874, Qatar.

Hua-Liang Wei is with the Department of Automatic Control and Systems Engineering, The University of Sheffield, Sheffield S1 3DJ, U.K (e-mail: w.hualiang@sheffield.ac.uk).

patterns from nonstationary EEG signals, provides a new approach to connectivity feature representation. Currently, the most commonly used approaches for dynamic GC analysis can be broadly categorized into three classes: sliding window method [17], adaptive multivariate estimation [18], and parametric modelling approach [19]. In the sliding window approach, the detection performance can be significantly affected by the choice of window size [20]. Most adaptive methods set fixed model structures and estimate model parameters based on recursive least squares (RLS) or Kalman filtering [21]; they cannot track rapid varying causalities because of the slow convergence speed. In contrast, the parametric approach employing basis function expansion scheme can provide better dynamic causal features with high temporal resolution [22]. In such a detection approach, the underlying time-varying models of signals are represented using multiwavelet basis functions with good approximation properties [23, 24], and an effective model structure decision algorithm like regularized orthogonal forward regression (ROFR) [25] is applied to reduce and refine the initial model; then both rapid and slow varying causalities between nonstationary signals can be successfully detected [22]. However, the pairwise time-domain GC approach proposed in [22] ignores frequency information and indirect effects caused by mutual sources, which are crucial in MI recognition due to the essential identity of different tasks is the specific regulation of a multi-channel EEG pattern in determined frequency ranges.

The classification method is an another vital part of a connectivity-based MI-BCI system; however, the advantages and potentials of the classifying algorithms for EEG classification have not been fully explored [26]. Recently, a prominent advance in machine learning is the application of deep learning with convolutional neural networks (ConvNets), and the capacity of ConvNets for MI-EEG decoding has also been investigated [27, 28]. Tabar *et al.* [29] studied a deep network combining ConvNet and stacked autoencoders to extract time-frequency features and classify MI-EEGs, where the classification results outperformed the classic filter bank CSP (FBCSP) algorithm. Nevertheless, there still exists important methodological problems on EEG analysis with ConvNets, including the large requirement on number of training data, the difficulty of interpretation, and the extremely unstable decoding performances across different participants.

In this paper, a novel multiwavelets-ROFR based boosted ConvNets with causality images is proposed for MI-EEG decoding. The proposed framework includes three key steps. First, a high-resolution time-frequency conditional Granger causality (TF-CGC) representation is developed by modifying the formulation of Geweke's spectral measure [30] with the time-varying autoregressive with exogenous input (TVARX) models of nonstationary signals. Second, the fundamental TVARX models for TF-CGC analysis are accurately identified using the multiwavelets-ROFR method, where the ROFR algorithm [25, 31] is used to determine the parsimonious model structure and associated parameters via a regularized loss function. Finally, the boosted ConvNets making use of high-resolution connectivity distributions is constructed to decode MI-EEG signals. With the employment of deep learning methods including cropping and boosting strategies, discriminative causal features can be extracted from the multi-domain (time, frequency and location) causality input images through spatial and temporal convolution. The performance of our proposed decoding scheme is evaluated on a publicly available MI-EEG dataset from BCI competition. Comparing to the state-of-the-art studies, better classification performance is obtained by the proposed framework. The main contribution of this work is that for the first time the deep ConvNet technique is introduced to explore single-trial time-frequency connectivity patterns for a robust and efficient MI classification. Additionally, the multiwavelets-ROFR modelling method, which can effectively identify TVARX models with good generalization property, is applied to TF-CGC analysis to achieve better TF causality distributions from multi-channel EEG data. As a result, the proposed multiwavelets causality-based boosted ConvNets decoding scheme provides a powerful solution to MI-EEG signal classification.

## II. METHODS

This study introduces a new deep ConvNet approach for MI-EEG decoding based on dynamic TF-CGC analysis. The proposed multiwavelets-ROFR method can produce high-resolution TF-CGC distributions from nonstationary multi-channel EEGs. The boosted ConvNets, which aims to extract multi-domain (time, frequency and location) discriminative features from causality images by spatial and temporal convolutions, is further designed for classifying EEG signals recorded from left- and right-hand MI tasks.

### A. High-resolution TF-CGC Analysis

To detect time-varying spectral causalities between any two channels out of three or more simultaneous nonstationary EEG signals, we first present a high-resolution TF-CGC analysis framework which generalizes the traditional Geweke spectral measure based on the parametric TVARX modelling approach. Consider three nonstationary processes $X = \{x(t)\}$, $Y = \{y(t)\}$ and $Z = \{z(t)\}$, with sampling index $t = 1, 2, \cdots N$, and the time-frequency causal influence from $Y$ to $X$ conditional on $Z$ expressed as $F_{Y \to X|Z}(t,f)$ is to be evaluated. Assume the bivariate TVARX representation of $x(t)$ and $z(t)$ is

$$x(t) = \sum_{k=1}^{K_1} a_{11,k}(t) x(t-k) + \sum_{k=1}^{K_2} a_{12,k}(t) z(t-k) + e_1(t)$$
$$z(t) = \sum_{k=1}^{K_1} a_{21,k}(t) x(t-k) + \sum_{k=1}^{K_2} a_{22,k}(t) z(t-k) + e_2(t)$$
(1)

where the initial noise terms $e_1(t)$ and $e_2(t)$ can be correlated with each other and their time-varying covariance matrix is $\Xi_1 = [(\Sigma_1(t)\ \Delta_1(t)), (\Delta_1(t)\ \Sigma_2(t))]^T$. Specifically $\Sigma_1(t) = \text{var}(e_1(t))$, $\Sigma_2(t) = \text{var}(e_2(t))$ and $\Delta_1(t) = \text{cov}(e_1(t), e_2(t))$ are calculated by a general recursive expression $\sigma(t+1) = (1-\zeta)\sigma(t) + \zeta u_1(t) u_2(t), 0 < \zeta < 1$ [32] with $\{u_1(t) = e_1(t), u_2(t) = e_1(t)\}, \{u_1(t) = e_2(t), u_2(t) = e_2(t)\}$ and $\{u_1(t) = e_1(t), u_2(t) = e_2(t)\}$, respectively. Define the lag operator g as $gx(t) = x(t-g)$, then (1) can be rewritten as

$$\begin{pmatrix} a_{11}(g) & a_{12}(g) \\ a_{21}(g) & a_{22}(g) \end{pmatrix} \begin{pmatrix} x(t) \\ z(t) \end{pmatrix} = \begin{pmatrix} e_1(t) \\ e_2(t) \end{pmatrix}$$
(2)

where $a_{11}(0) = a_{22}(0) = 1$, $a_{12}(0) = a_{21}(0) = 0$. The independence of $e_1(t)$ and $e_2(t)$ is necessary for the definition of spectral domain causality, thus the normalization procedure introduced by Geweke is used to remove the correlation and make the identification of an intrinsic part and a causal part possible in time-varying cases [30]. The transformation consists of left-multiplying $C(t) = [(1\ 0), (-\Delta_1(t)/\Sigma_1(t)\ 1)]^T$ on both sides of (2) at each time index [33], and the resulting normalized form is given as

$$\begin{pmatrix} A_{11}(g) & A_{12}(g) \\ A_{21}(g) & A_{22}(g) \end{pmatrix} \begin{pmatrix} x(t) \\ z(t) \end{pmatrix} = \begin{pmatrix} \varepsilon_1(t) \\ \varepsilon_2(t) \end{pmatrix} \quad (3)$$

where $A_{11}(0) = A_{22}(0) = 1$, $A_{12}(0) = 0$, $A_{21}(0)$ is generally not zero, $\text{cov}(\varepsilon_1(t), \varepsilon_2(t)) = 0$, and note that $\text{var}(\varepsilon_1(t)) = \Sigma_1(t)$. Taking time-frequency transform of both sides of (3) yields

$$\overbrace{\begin{pmatrix} A_{11}(t,f) & A_{12}(t,f) \\ A_{21}(t,f) & A_{22}(t,f) \end{pmatrix}}^{A(t,f)} \begin{pmatrix} X(t,f) \\ Z(t,f) \end{pmatrix} = \begin{pmatrix} \mathrm{E}_1(t,f) \\ \mathrm{E}_2(t,f) \end{pmatrix} \quad (4)$$

where the components of the coefficient matrix $A(t,f)$ are $A_{11}(t,f) = 1 - \sum_{k=1}^{K_1} A_{11,k} e^{-i2\pi kf/f_s}$, $A_{12}(t,f) = -\sum_{k=1}^{K_2} A_{12,k} e^{-i2\pi kf/f_s}$, $A_{21}(t,f) = -\sum_{k=1}^{K_1} A_{21,k} e^{-i2\pi kf/f_s}$, $A_{22}(t,f) = 1 - \sum_{k=1}^{K_2} A_{22,k} e^{-i2\pi kf/f_s}$ with $i = \sqrt{-1}$ and $f_s$ being the sampling frequency.

Let the trivariate TVARX models of all three processes $x(t)$, $y(t)$ and $z(t)$ be

$$x(t) = \sum_{k=1}^{K_1} b_{11,k}(t) x(t-k) + \sum_{k=1}^{K_2} b_{12,k}(t) y(t-k) + \sum_{k=1}^{K_3} b_{13,k}(t) z(t-k) + e_3(t)$$

$$y(t) = \sum_{k=1}^{K_1} b_{21,k}(t) x(t-k) + \sum_{k=1}^{K_2} b_{22,k}(t) y(t-k) + \sum_{k=1}^{K_3} b_{23,k}(t) z(t-k) + e_4(t)$$

$$z(t) = \sum_{k=1}^{K_1} b_{31,k}(t) x(t-k) + \sum_{k=1}^{K_2} b_{32,k}(t) y(t-k) + \sum_{k=1}^{K_3} b_{33,k}(t) z(t-k) + e_5(t)$$

(5)

where the time-varying covariance matrix of the noise terms can be computed with the recursive formula similarly as $\Xi_1$, and is denoted as $\Xi_2 = \left[\left(\Sigma_{xx}(t), \Sigma_{xy}(t), \Sigma_{xz}(t)\right), \left(\Sigma_{yx}(t), \Sigma_{yy}(t), \Sigma_{yz}(t)\right), \left(\Sigma_{zx}(t), \Sigma_{zy}(t), \Sigma_{zz}(t)\right)\right]^T$. For (5), the normalization process involves left-multiplying both sides by the time-varying matrix $D(t) = D_2(t) \cdot D_1(t)$, where $D_1(t) = \left[(1,0,0), \left(-\Sigma_{yx}(t)\Sigma_{xx}^{-1}(t), 1, 0\right), \left(-\Sigma_{zx}(t)\Sigma_{xx}^{-1}(t), 0, 1\right)\right]^T$, $D_2(t) = \left[(1,0,0), (0,1,0), \left(0, -\left(\Sigma_{zy}(t) - \Sigma_{zx}(t)\Sigma_{xx}^{-1}(t)\Sigma_{xy}(t)\right)\left(\Sigma_{yy}(t) - \Sigma_{yx}(t)\Sigma_{xx}^{-1}(t)\Sigma_{xy}(t)\right)^{-1}, 1\right)\right]^T$ [33]. The associated normalized equations are

$$\begin{pmatrix} B_{11}(g) & B_{12}(g) & B_{13}(g) \\ B_{21}(g) & B_{22}(g) & B_{23}(g) \\ B_{31}(g) & B_{32}(g) & B_{33}(g) \end{pmatrix} \begin{pmatrix} x(t) \\ y(t) \\ z(t) \end{pmatrix} = \begin{pmatrix} \varepsilon_3(t) \\ \varepsilon_4(t) \\ \varepsilon_5(t) \end{pmatrix} \quad (6)$$

where the noise terms $\varepsilon_3(t)$, $\varepsilon_4(t)$, $\varepsilon_5(t)$ are now independent, and their time-varying variances are $\tilde{\Sigma}_{xx}(t)$, $\tilde{\Sigma}_{yy}(t)$ and $\tilde{\Sigma}_{zz}(t)$, respectively. Also, similarly to (4), the time-varying spectral decomposition of (6) can be calculated and expressed as

$$\overbrace{\begin{pmatrix} B_{11}(t,f) & B_{12}(t,f) & B_{13}(t,f) \\ B_{21}(t,f) & B_{22}(t,f) & B_{23}(t,f) \\ B_{31}(t,f) & B_{32}(t,f) & B_{33}(t,f) \end{pmatrix}}^{B(t,f)} \begin{pmatrix} X(t,f) \\ Y(t,f) \\ Z(t,f) \end{pmatrix} = \begin{pmatrix} \mathrm{E}_3(t,f) \\ \mathrm{E}_4(t,f) \\ \mathrm{E}_5(t,f) \end{pmatrix} \quad (7)$$

According to the key relationship of conditional causality in time and frequency domain [30], the problem of measuring time-dependent spectral causal connectivity from $Y$ to $X$ conditional on $Z$ can thus be converted into the calculation of causal influence from $Y\varepsilon_2$ to $\varepsilon_1$. In order to obtain $F_{Y\varepsilon_2 \to \varepsilon_1}(t,f)$, the variance of $\varepsilon_1$ is decomposed over time and frequency. By rearranging (4) and (7) into the transfer function format, we get

$$\begin{pmatrix} X(t,f) \\ Z(t,f) \end{pmatrix} = \overbrace{\begin{pmatrix} G_{xx}(t,f) & G_{xz}(t,f) \\ G_{zx}(t,f) & G_{zz}(t,f) \end{pmatrix}}^{G(t,f)} \begin{pmatrix} \mathrm{E}_1(t,f) \\ \mathrm{E}_2(t,f) \end{pmatrix} \quad (8)$$

$$\begin{pmatrix} X(t,f) \\ Y(t,f) \\ Z(t,f) \end{pmatrix} = \overbrace{\begin{pmatrix} H_{xx}(t,f) & H_{xy}(t,f) & H_{xz}(t,f) \\ H_{yx}(t,f) & H_{yy}(t,f) & H_{yz}(t,f) \\ H_{zx}(t,f) & H_{zy}(t,f) & H_{zz}(t,f) \end{pmatrix}}^{H(t,f)} \begin{pmatrix} \mathrm{E}_3(t,f) \\ \mathrm{E}_4(t,f) \\ \mathrm{E}_5(t,f) \end{pmatrix} \quad (9)$$

where the TF domain transfer function $G(t,f)$ and $H(t,f)$ are the inverse of the coefficient matrixes in (4) and (7), i.e. $A(t,f)$ and $B(t,f)$, represented as $G(t,f) = A^{-1}(t,f)$ and $H(t,f) = B^{-1}(t,f)$, respectively.

Assume that $X(t,f)$ and $Z(t,f)$ in (6) and (7) are identical [33], then the two equations can be combined to yield

$$\begin{pmatrix} \mathrm{E}_1(t,f) \\ Y(t,f) \\ \mathrm{E}_2(t,f) \end{pmatrix} = \overbrace{\begin{pmatrix} \Re_{xx}(t,f) & \Re_{xy}(t,f) & \Re_{xz}(t,f) \\ \Re_{yx}(t,f) & \Re_{yy}(t,f) & \Re_{yz}(t,f) \\ \Re_{zx}(t,f) & \Re_{zy}(t,f) & \Re_{zz}(t,f) \end{pmatrix}}^{\Re(t,f)} \begin{pmatrix} \mathrm{E}_3(t,f) \\ \mathrm{E}_4(t,f) \\ \mathrm{E}_5(t,f) \end{pmatrix} \quad (10)$$

where $\Re(t,f) = G^{-1}(t,f)H(t,f)$. The time-dependent spectrum of $\mathrm{E}_1$ can then be decomposed into the following three parts based on (10)

$$S_{\mathrm{E}_1}(t,f) = \Re_{xx}(t,f)\tilde{\Sigma}_{xx}(t)\Re_{xx}^*(t,f) + \Re_{xy}(t,f)\tilde{\Sigma}_{yy}(t)\Re_{xy}^*(t,f) + \Re_{xz}(t,f)\tilde{\Sigma}_{zz}(t)\Re_{xz}^*(t,f) \quad (11)$$

where the upper script '*' denotes complex conjugate and transpose of a matrix. Note that the first term can be regarded as the intrinsic power and the remaining two terms represent the combined causal relations from $Y$ and $\varepsilon_2$. Hence the causality from $Y\varepsilon_2$ to $\varepsilon_1$, also the expression for time-varying spectral Granger causality from $Y$ to $X$ conditional on $Z$ is

$$F_{Y \to X|Z}(t,f) = F_{Y\varepsilon_2 \to \varepsilon_1}(t,f) = \ln \frac{\left|S_{\mathrm{E}_1}(t,f)\right|}{\left|\Re_{xx}(t,f)\tilde{\Sigma}_{xx}(t)\Re_{xx}^*(t,f)\right|} \quad (12)$$

The spectral function in (12) is a continuous function of frequency $f$, and can be used to measure the spectral causality at any desired frequency from 0 up to the Nyquist frequency $f_s/2$. Generally the frequency resolution is not infinite, but depends on the corresponding parameter estimates and the underlying model order. The calculated value of $F_{Y \to X|Z}(t,f)$ represents the strength of interaction between the input and the output of a multivariate system, in time and frequency domain. However, the reliability of the identified TVARX model is also affected by the model structure and the number of sampling data. Thus, a hypothesis test is required to decide whether the detected influence in the sampled data is statistically significant. The thresholds for statistical significance are computed from surrogate data by a permutation procedure under a null hypothesis of no interdependence at the significance level $p < 10^{-6}$. Note that the aforementioned theoretical formulations, developed the case of three signals, can be extended to four and more time series.

*B. Identification of TVARX Models based on Multiwavelets and ROFR*

From the above descriptions, in order to obtain high-resolution TF-CGC representations from multivariate coupling systems, the accurate identification of TVARX models for characterizing nonstationary signals plays a key role. In this study, the multiwavelets expansion method and ROFR algorithm are applied to efficiently estimate time-varying models. Particularly, for the TVARX model given in (1), with $x(t)$ and $z(t)$ being output and input, respectively, can be expressed by

$$x(t) = \sum_{k=1}^{K_1} a_{11,k}(t)x(t-k) + \sum_{k=1}^{K_2} a_{12,k}(t)z(t-k) + e_1(t) \\ = \phi^T(t)\vartheta(t) + e_1(t) \quad (13)$$

where $K_1$ and $K_2$ are the maximum time lags of $x(t)$ and $z(t)$, respectively, $\phi(t) = [x(t-1), \cdots, x(t-K_1), z(t-1), \cdots, z(t-K_2)]^T$ denote monomials of delayed output and input terms, $\vartheta(t) = [a_{11,1}(t), \cdots, a_{11,K_1}(t), a_{12,1}(t), a_{12,K_2}(t)]^T$ is a time-varying parameter vector to be estimated, and $e_1(t)$ is an independent zero mean noise sequence. A basis function expansion method [34] is adopted to identify time-varying models in this work. Multiwavelet basis functions have been proved effective for tracking both rapid and smooth parameter variations in time-varying processes [24, 35], so multiwavelets are used as the building blocks for model parameter approximation. The time-varying parameters in (13) can be expanded using multiple wavelet basis functions as below

$$a_{11,k}(t) = \sum_s \sum_{l \in \Gamma_s} c_{k,l}^s \varphi_{l,j}^s\left(\frac{t}{N}\right) \\ a_{12,k}(t) = \sum_s \sum_{l \in \Gamma_s} d_{k,l}^s \varphi_{l,j}^s\left(\frac{t}{N}\right) \quad (14)$$

where $\varphi_{l,j}^s(u) = 2^{j/2}\varphi^s(2^j u - l)$ are shifted and dilated versions of a wavelet basis function $\varphi^s(u)$ with shift indices $l \in \Gamma_s, \Gamma_s = \{l: -s \leq l \leq 2^j - 1\}$ and wavelet scale $j$, $s$ denotes the order of basis functions, $c_{k,l}^s$ and $d_{k,l}^s$ are time invariant expansion parameters, and the function variable $u = t/N$ is normalized within $[0,1]$. The initial time-varying modelling problem then becomes time invariant because $c_{k,l}^s$ and $d_{k,l}^s$ are now time-invariant.

Cardinal B-splines are an important class of basis functions with excellent properties such as compactly supported, analytically formulated and multiresolution analysis oriented, making them appropriate for constructing multi-resolution wavelet decompositions, and enable the operation of decomposition to be more convenient [36]. Taking the cardinal B-splines as the basis functions, the $\varphi_{l,j}^s(\cdot)$ can be represented by the $s$-th order B-spline $\beta_s$ as $\varphi_{l,j}^s(u) = 2^{j/2}\beta_s(2^j u - l)$, where $l, j$ denote the shifted and dilated versions of wavelet $\beta_s$. Generally, letting $j$ be 3 or a larger number is appropriate for many B-splines applications, and a practical selection of the wavelets are $\{\varphi_{l,j}^s: s = 3,4,5\}$ [24]. Detailed descriptions of B-splines can be found in [37]. By expanding the time-varying parameters with multiple B-spline basis functions, the TVARX model (13) becomes

$$x(t) = \sum_{k=1}^{K_1} \sum_s \sum_{l \in \Gamma_s} c_{k,l}^s \varphi_{l,j}^s\left(\frac{t}{N}\right)x(t-k) + \sum_{k=1}^{K_2} \sum_s \sum_{l \in \Gamma_s} d_{k,l}^s \varphi_{l,j}^s\left(\frac{t}{N}\right)z(t-k) + e_1(t) \\ = \psi^T(t)\theta + e_1(t) \quad (15)$$

where $\psi(t) = [x(t-1)\boldsymbol{\varphi}(t), \cdots, x(t-K_1)\boldsymbol{\varphi}(t), z(t-1)\boldsymbol{\varphi}(t), \cdots, z(t-K_2)\boldsymbol{\varphi}(t)]^T$ with $\boldsymbol{\varphi}(t) = \varphi_{l,j}^s\left(\frac{t}{N}\right)$ is the expanded regressor vector at time $t$ and $\theta = [c_{1,l}^s, \cdots, c_{K_1,l}^s, d_{1,l}^s, \cdots, d_{K_2,l}^s]^T$ is the corresponding time-invariant parameter vector. Equation (15) indicates that the basis function expansion method converts the identification of a time-varying model to solving a time-invariant regression problem. However, the number of candidate regressors (model terms) in (15) can be very large, resulting in potentially high redundancy in the model and the difficulty to estimate the model parameters. Therefore selecting significant terms and detecting the correct parsimonious model structure are a crucial task for the model identification.

Denote the resulting matrix generated by all the vectors $\psi(t)$ in (15) at $N$ discrete time samples by $\Psi$ and the associated time-invariant parameter vector by $\Theta$, then the corresponding compact regression matrix form is $X = \Psi\Theta + e$, where $X = [x(1), \cdots, x(N)]^T$ is the output vector, $\Psi = [\psi^T(1), \cdots, \psi^T(N)]^T$ is a regressor matrix, and $e = [e_1(1), \cdots, e_1(N)]^T$ is the approximation error. Each of the columns of the matrix $\Psi$ represents a vector matching a candidate model term specified in the dictionary $W = \{\xi_1, \xi_2, \cdots, \xi_M\}$, and the model structure detection problem is to determine significant basis vectors from $\Psi$, which is equivalent to selecting a subset of model terms $W_q = \{\xi_{L_1}, \xi_{L_2}, \cdots, \xi_{L_q}\}$ ($q \ll M$) from the candidate set $W$. Thus the output $X$ can be approximated by a linear combination of the selected terms as $X = \xi_{L_1}\pi_1 + \xi_{L_2}\pi_2 + \cdots + \xi_{L_q}\pi_q + e$ or in the matrix form $X = \Phi\Pi + e$, where the regression matrix $\Phi = [\xi_{L_1}, \xi_{L_2}, \cdots, \xi_{L_q}]$ is full column rank and $\Pi = [\pi_1, \pi_2, \cdots, \pi_q]^T$ is the parameter vector.

In this study, the zero-order ROFR algorithm [25, 31], which is developed based upon the well-known orthogonal forward regression (OFR) method [35, 38] by applying the zero-order regularization, is used for model structure selection and model reduction. The significant terms can be selected based on the regularized error reduction ratio (RERR) defined as [25]

$$RERR(X,\xi) = \frac{\langle X,\xi\rangle^2}{X^T X(\langle\xi,\xi\rangle + \rho)} \quad (16)$$

where $\rho \geq 0$ is the regularization parameter, $X$ is the output vector, $\xi$ is a candidate term, and the symbol $\langle\cdot,\cdot\rangle$ denotes the inner product of two vectors. The pseudocode for the ROFR algorithm is presented in Appendix (Algorithm 1), and the detailed selection procedure can be found from there.

Furthermore, the penalized error-to-signal ratio (PESR) criterion [19, 39] shown below is adopted to determine the model size

$$PESR(\alpha) = \frac{1}{(1-\mu\alpha/N)^2}\left(1 - \sum_{m=1}^{\alpha} RERR_{L_m}\right) \quad (17)$$

where $\alpha$ is the number of selected terms, $N$ is the length of sampled data, and the adjustable parameter $\mu$ is suggested to be chosen between 5 and 10 [19, 39]. The regularized term search procedure terminates if $PESR(\alpha)$ reaches the minimum at $\alpha = q$, and yields a $q$-term model. Orthogonally decomposing the selected regression matrix $\Phi$ (which is full rank in columns) as $\Phi = QV$, where $Q$ is a $N \times q$ matrix with orthogonal columns and $V$ is an $q \times q$ unit upper triangular matrix. Then the associated coefficient vector $\Pi$ can be calculated by $V\Pi = K$ with $K = (Q^T Q)^{-1}V^T X$, and finally the time-varying parameters in the TVARX model (13) can be approximated using the resultant estimates. Similar to the model (13), other time-varying processes in (1) and (5) for TF-CGC analysis can also be identified by applying the proposed multiwavelets and ROFR method. The significant TF-CGC values are further obtained through (12) to form a high-resolution time-frequency connectivity distribution for the classification of MI-EEGs.

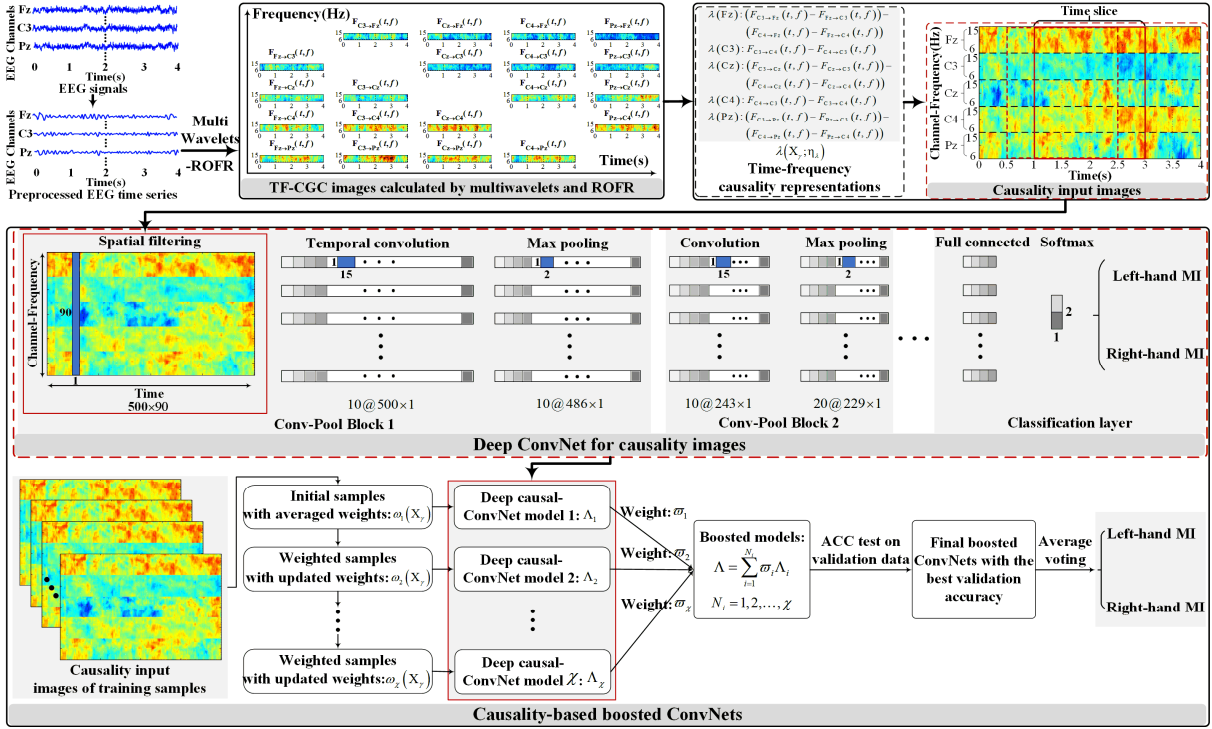

Fig. 1. Flowchart of proposed decoding framework.

*C. EEG Decoding using Boosted ConvNets with Causality Images*

*1) Overview of the decoding framework*

An EEG decoding network based on TF-CGC images is introduced in this section. Denote EEG data from one subject as $\{(X_1, y_1), \cdots, (X_Y, y_Y)\}$, where $Y$ is the number of recorded trials per subject. The input data $\{X_\gamma \in \mathbb{R}^{I.T} | 1 \leq \gamma \leq Y\}$ represents the preprocessed signals of $I$ electrodes and $T$-discretized time steps recorded per trial. $\{y_\gamma \in \delta | 1 \leq \gamma \leq Y\}$ is the corresponding class label of trial $\gamma$ which takes values from a set of class labels $\delta = \{$"Left-hand", "right-hand"$\}$. The decoder is trained on the existing trials to assign the label $y_\gamma$ to trail $X_\gamma$ using the output of a parametric classifier $f(X_\gamma; \eta): \mathbb{R}^{I.T} \to \delta$ with parameters $\eta$. The classifier $f(X_\gamma; \eta)$ of the proposed decoding scheme can be decomposed into two parts: the first part that calculates the time-frequency causality representation (images) $\lambda(X_\gamma; \eta_\lambda)$ with parameters $\eta_\lambda$; and the second part consisting of ConvNets $\Lambda$ that are designed for causal feature extraction and further classification using causality images, that is $f(X_\gamma; \eta) = \Lambda(\lambda(X_\gamma; \eta_\lambda); \eta_\Lambda)$. The causality-based decoding process is graphically shown in Fig. 1. To be specific, the multi-domain (temporal, spectral and spatial) causality input images are first obtained by combining multichannel TF-CGC distributions which are computed via multiwavelets-ROFR from preprocessed EEG time series. The TF-CGC calculation process is detailedly illustrated in Appendix (Fig. 2). Then the deep ConvNets using boosting strategies are performed to extract discriminating causal features and classify MI-EEG signals. The details of the proposed framework are discussed in the following three parts.

*2) Representation of causality input images based on high-resolution TF-CGC analysis*

Given that *alpha* band (8-14 Hz) is one of the most primarily studied frequency bands when investigating the oscillatory cortical activity during motor operations [6], the multiwavelets-ROFR TF-CGC method is performed in *alpha* band to create the image-based representation of MI-EEGs, where the connectivity patterns distributed at different frequencies as well as time and location are preserved. Particularly, we consider the frequency range 6-15 Hz to represent *alpha* band, which fully cover the whole band. This is slightly different from literature but can result in a better data representation in our experiments. The frequency resolution of the TF-CGC analysis is set as $1/10 Hz$, which is adequate for describing spectral causal information of EEG in this decoding scheme [40]. Let $t_s$ be the time length analyzed for each trial, and assume that signals are sampled with a period of $1/250 s$ (i.e. sampling frequency $f_s = 250 Hz$), then the TF-CGC decomposition leads to a $(t_s \times 250) \times 90$ image for each channel-pair, where $t_s \times 250$ and $90 = 9 \times 10$ ($9 = 15 Hz - 6 Hz$) are the number of samples along time and frequency axes, respectively.

To reduce the computational cost caused by TVARX modelling in multichannel causality analysis and also lower the dimension of the 2D input image in frequency (the dimension is $N_C \times 90$ with $N_C$ denoting the number of adopted electrodes) [28], only EEG signals recorded from 5 most commonly studied electrodes in MI related researches (Fz, C3, Cz, C4, Pz) are used for classification in this work ($N_C = 5$) [41]. A total of 20 TF-CGC images are obtained from these five electrodes. Based upon the MI-GC results given in [41, 42], we denote
$\{(F_{C3 \to Fz}(t,f) - F_{Fz \to C3}(t,f)) - (F_{C4 \to Fz}(t,f) - F_{Fz \to C4}(t,f))\}$,
$\{F_{C3 \to C4}(t,f) - F_{C4 \to C3}(t,f)\}$, $\{(F_{C3 \to Cz}(t,f) - F_{Cz \to C3}(t,f)) - (F_{C4 \to Cz}(t,f) - F_{Cz \to C4}(t,f))\}$, $\{F_{C4 \to C3}(t,f) - F_{C3 \to C4}(t,f)\}$,
$\{(F_{C3 \to Pz}(t,f) - F_{Pz \to C3}(t,f)) - (F_{C4 \to Pz}(t,f) - F_{Pz \to C4}(t,f))\}$
as the causal representations of Fz, C3, Cz, C4 and Pz, respectively, aiming to increase description difference between MI categories and simplify the network input format by taking full advantage of these causal images, where $F_{c1 \to c2}(t,f)$ expresses the TF-CGC from channel $c1 \in \{Fz, C3, Cz, C4, Pz\}$ to $c2 \in \{Fz, C3, Cz, C4, Pz\} \backslash c1$ conditional on the other three

EEG channels. The time-frequency relations are then combined in a way that further exploits the electrode neighboring information (Fig. 1), and the consequent multi-domain (time, frequency and location) causality image is a 2D array with the size of $(t_s \times 250) \times 450$, where $t_s \times 250$ is the width (horizontal i.e. time axis), $450 = N_C \times 90$, with $N_C = 5$, is the height (vertical i.e. frequency axis). The high dimensionality of the images in frequency (450 samples), and the presence of artifacts and noises, make it challenging to design an ideal deep learning framework for EEG classification. Since the causal message contained in spectrum from multichannel is abundant enough for classification, the causality images in frequency is thereby average down-sampled to 90 data without significant information loss. These causality images produced from MI trial samples are employed as the network inputs for constructing the next causality-based ConvNets.

*3) Deep ConvNet for causality images*

A deep ConvNet architecture, inspired by the successful application in computer vision [43], is designed to recognize TF-CGC images and further decode MI-EEG signals. The ConvNet is a multi-layer neural network comprised of three categories of components: the convolutional layer, the pooling layer, and the full connected layer. The input causality image is convolved with a set of filters in the convolutional layer, and the causal features in time, spectrum and space domain are extracted by a nonlinear transformation and subsampled to a smaller size in the pooling layer. A back-propagation algorithm is used to train network weights aiming at decreasing the classification error.

A number of convolution-max-pooling blocks are applied in the proposed causality-based deep ConvNet, in which the first block is specially designed to better handle the high dimensionality of input causality images, followed by standard convolution-max-pooling blocks together with a softmax classification layer. The first convolutional block (also named spatio-temporal block) is split into a spatial filtering layer and a temporal convolution layer with no activation function in between. In the first spatial layer, 1D filters with the same height as the input $(1 \times 90)$ is employed along the horizontal (time) axis to extract spectrum and location causal features; and in the second temporal layer, each filter of size $\tau \times 1$ executes a 1D convolution over time. Reducing the number of parameters trained in the ConvNet properly would be useful to get a well-performed model under condition of limited number of training data, thus the number of filters in $\kappa$-th blocks $P_\kappa$ ($\kappa = 2,3,\cdots$) satisfies $P_\kappa = 2P_{\kappa-1}$, and the kernel size of the following convolution layers is also $\tau \times 1$. The exponential linear units (ELUs) defined in [44] are selected as the activation functions. Sampling factors are set to 2 in the max-pooling layers. The last full connected softmax layer includes two outputs denoting left- and right-hand MI tasks. In addition, batch normalization [45] is used to the output of convolutional layers before the nonlinear activation, and the inputs to all convolutional layers except the first are dropped out with a probability of 0.5. In the current model, the network parameters need to be selected by performing a ten-fold cross validation include kernel size $\tau \times 1$ ($\tau$ ranges from 10 to 20), kernel number of the first convolutional layer $P_1$ (ranges from 5 to 30 with a step of 5), and the number of convolution-pooling blocks (ranges from 1 to 5). Particularly, the causality-based deep ConvNet shown in Fig. 1 is a sample visualization view which uses $\tau_1 = 15, P_1 = 10$ as an example.

As for the problem of the network training, given that multiple crops of the input image in ConvNets is an effective approach to increase classification accuracy for object recognition [46], the training strategy using crops is adopted in this work, where sliding input windows along the time axis are introduced to get more training samples for the network. Formally, for a 4 s original trial $X_\gamma \in \mathbb{R}^{I,T}$ with $I$ electrodes and $T$ timesteps, a set of crops with crop size 2 s are generated as timeslices of the trial: $\aleph_\gamma = \{X_\gamma^{1\cdots I, t\cdots t+(f_s \times 2s)} | t \in 1 \cdots T - (f_s \times 2s)\}$. All of these crops are new training data samples for the decoder and will get that same label $y_\gamma$ as the original trial. We collect crops starting from the trial start to the trial end with sampling interval of 0.5 s, i.e. $\aleph_\gamma = \{X_\gamma^{1\cdots I, t\cdots t+(f_s \times 2s)} | t \in 1, 1 + (f_s \times 0.5s), 1 + (f_s \times 1s), 1 + (f_s \times 1.5s), 1 + (f_s \times 2s)\}$. Overall, this results in 5 crops and corresponding 5 label predictions per trial. Moreover, note that the ConvNet input size in time axis will be 500 samples for 250 Hz sampling rate using crops ($t_s$ = cropping size 2s), thus the dimension of the causality input images is $500 \times 90$. The large number of parameters in the ConvNet are optimized by minimizing the categorical cross-entropy using Adam [47] together with an early stopping method [43]. The early stopping strategy divides the training set into a training (80%) and validation (20%) fold, and the training stops when the validation accuracy does not increase for 500 epochs.

*4) Causality-based boosted ConvNets using AdaBoost*

There is a degree of correlation among the cropped training examples, and the importance of the samples during different time slices may be disparate for MI classification as the responses to MI tasks are primarily existing in the previous 2 s after the trial starts. In order to further mine and utilize the information from the augmented samples, the augmented training samples are re-weighted using the adaptive boosting (AdaBoost) algorithm which is realized by iteratively changing the sample data distribution [48]. The core idea of AdaBoost is to use a set of general classifiers by a certain method of cascade to form a strong classifier. The deep ConvNet for causality images described above is applied as the basic classifier that needs to be boosted. Assume that $\chi$ iterations are automatically generated in the boosting process, then the joint model with best validation accuracy is chosen as the final classification model. The value of $\chi$ is initially set to be large enough, however, the results show that the adjustment of samples is sufficient for improving the cascaded model performance when $\chi$ arrives at around 20, in which case the computation load of the boosted model is easily affordable due to the small number of required iterations. Additionally, similar to the cropped training samples, each trial of the testing set (4 s) is split into 5 parts (with each size 2 s and sampling interval 0.5 s) as timeslices of the trial from the start of the trial to the end. These 5 testing samples during different time periods are predicted by the deep boosted model independently, and the average voting of the 5 prediction labels determines the final label of the original testing trial. The pseudocode for the boosted ConvNets is given in Appendix (Algorithm 2), and the schematic diagram of the integrated testing framework using causality-based boosted ConvNets is shown in Fig. 1.

III. EXPERIMENTAL RESULTS

*A. Data and preprocessing*

In this study, the effectiveness of the proposed causality-based boosted ConvNets is evaluated on the BCI Competition

> REPLACE THIS LINE WITH YOUR PAPER IDENTIFICATION NUMBER (DOUBLE-CLICK HERE TO EDIT) <       7IV Dataset-IIa [49], where a binary classification is performed. Specifically, the EEG data were recorded from nine subjects performing four different MI tasks, consisted of two sessions carried out on different days. Each session contains 72 four-second trials per MI task. The EEG signals were acquired by 22 Ag/AgCl electrodes and sampled at 250 Hz. Here, two MI tasks, i.e. imagined movement of left- and right-hand, are selected for the binary classification. Thus the training set involves the 144 trials of the first session, and the testing set refers to the 144 trials of the second session.

As described in the previous section, the frequency range 6-15 Hz (covering *alpha* band) is used for data representation in this causality-based EEG decoding scheme. To obtain 6-15 Hz EEG signals, the noise-assisted multivariate empirical mode decomposition (NA-MEMD) algorithm [50] is adopted, where the 22-channel EEG data are decomposed with two additional noise channels (SNR = 20dB, SNR = 40dB). The intrinsic mode functions (IMFs) prepared for the subsequent decoding process are then determined based on the Hilbert-Huang spectrum of each yielded IMF, where the ones most relevant to *alpha* rhythm are retained.

### B. Overview of classification performance

The filter bank common spatial patterns (FBCSP) algorithm [9], which was the best-performing method for the BCI competition dataset, is chosen as an adequate benchmark algorithm for the performance evaluation. Furthermore, ConvNets using different training strategies with the identical architecture for causality images, are also tested to show the effectiveness of the proposed deep boosted ConvNets. Specifically, for the basic causality-based ConvNet which does not employ cropping and boosting strategies, the single trial samples with the whole duration of 0-4 s are used for training. By contrast, 5 times more training data to the original training trials are generated in the ConvNet using crops. For the proposed boosted network, ConvNets with causality images using crops are further integrated by Adaboost algorithm (described in Fig. 1), and the average voting methods are introduced for the testing process. In these experiments, all methods are trained in a subject-independent way with trials of a single subject.

In general, the decoding performance can be evaluated by statistical measurements of sensitivity (SEN), specificity (SPE), accuracy (ACC) [40] and kappa [29]. The classification results achieved on the test data of each subject, and the associated mean and inter subject standard deviation (SD) of all subjects, obtained by each of the compared methods are show in Table 1.

From Table 1, although there exits variability in classification performance over subjects, overall the proposed method clearly outperforms the other methods in terms of sensitivity, specificity, accuracy and kappa value on average. In particular, mean sensitivity and specificity of all subjects obtained from the proposed approach are 84.10% and 85.34%, respectively; the higher value indicates that an effective discrimination between left and right-hand EEG signals can be achieved. Moreover, the proposed method produces an average classification accuracy of 84.72% and kappa of 0.6944, with corresponding standard deviation 3.18% and 0.0637, indicating the good robustness and better classification performance of our multiwavelets-ROFR based boosted network method than the FBCSP and other listed methods. As for the classifications derived for each individual, the proposed approach reaches the best accuracy and kappa for 7 out of 9 subjects. These experiment results demonstrate the efficiency of the proposed ConvNets with time-frequency causal images method, and is capable of decoding MI-EEG signals.

TABLE I
BINARY CLASSIFICATION RESULTS ON BCI COMPETITION IV DATASET-IIA

| Subject | Method | SEN (%) | SPE (%) | ACC (%) | kappa |
|---|---|---|---|---|---|
| S1 | FBCSP | **97.22** | 68.06 | 82.64 | 0.6528 |
|  | ConvNet | 86.11 | 73.61 | 79.86 | 0.5972 |
|  | ConvNet using corps | 59.72 | **98.61** | 79.17 | 0.5833 |
|  | **Boosted ConvNets using corps** | 77.78 | 93.06 | **85.42** | **0.7083** |
| S2 | FBCSP | 69.44 | 45.83 | 57.64 | 0.1528 |
|  | ConvNet | **79.17** | 70.83 | 75.00 | 0.5000 |
|  | ConvNet using corps | 70.83 | 80.56 | 75.69 | 0.5139 |
|  | **Boosted ConvNets using corps** | 75.00 | **84.72** | 79.86 | 0.5972 |
| S3 | **FBCSP** | 94.44 | **87.50** | 90.97 | 0.8194 |
|  | ConvNet | 76.39 | 76.39 | 76.39 | 0.5278 |
|  | ConvNet using corps | 91.67 | 77.78 | 84.72 | 0.6944 |
|  | Boosted ConvNets using corps | **97.22** | 81.94 | **89.58** | **0.7917** |
| S4 | FBCSP | 62.50 | 72.22 | 67.36 | 0.3472 |
|  | ConvNet | **87.50** | 70.83 | 79.17 | 0.5833 |
|  | ConvNet using corps | 68.06 | 80.56 | 74.31 | 0.4861 |
|  | **Boosted ConvNets using corps** | 76.39 | **84.72** | **80.56** | **0.6111** |
| S5 | FBCSP | 58.33 | **88.89** | 73.61 | 0.4722 |
|  | ConvNet | 62.50 | 80.56 | 71.53 | 0.4306 |
|  | ConvNet using corps | 97.22 | 63.89 | 80.56 | 0.6111 |
|  | **Boosted ConvNets using corps** | **100.00** | 72.22 | **86.11** | **0.7222** |
| S6 | FBCSP | 38.89 | 80.56 | 59.72 | 0.1944 |
|  | ConvNet | 80.56 | 80.56 | 80.56 | 0.6111 |
|  | ConvNet using corps | **93.06** | 70.83 | 81.94 | 0.6389 |
|  | **Boosted ConvNets using corps** | 77.78 | **87.50** | **82.64** | **0.6528** |
| S7 | FBCSP | 68.06 | 97.22 | 82.64 | 0.6528 |
|  | ConvNet | **97.22** | 51.39 | 74.31 | 0.4861 |
|  | ConvNet using corps | 62.50 | **100.00** | 81.25 | 0.6250 |
|  | **Boosted ConvNets using corps** | 83.33 | 86.11 | **84.72** | **0.6944** |
| S8 | **FBCSP** | 90.28 | 90.28 | 90.28 | 0.8056 |
|  | ConvNet | **95.83** | 69.44 | 82.64 | 0.6528 |
|  | ConvNet using corps | 76.39 | 88.89 | 82.64 | 0.6528 |
|  | Boosted ConvNets using corps | 84.72 | 87.50 | 86.11 | 0.7222 |
| S9 | FBCSP | **94.44** | 55.56 | 75.00 | 0.5000 |
|  | ConvNet | 86.11 | 75.00 | 80.56 | 0.6111 |
|  | ConvNet using corps | 84.72 | 76.39 | 80.56 | 0.6111 |
|  | **Boosted ConvNets using corps** | 84.72 | **90.28** | **87.50** | **0.7500** |
| Mean (SD) | FBCSP | 74.84 | 76.24 | 75.54 (12.24) | 0.5108 (0.2447) |
|  | ConvNet | 83.49 | 72.07 | 77.78 (3.64) | 0.5556 (0.0728) |
|  | ConvNet using corps | 78.24 | 81.95 | 80.09 (3.29) | 0.6018 (0.0659) |
|  | **Boosted ConvNets using corps** | 84.10 | 85.34 | 84.72 (3.18) | 0.6944 (0.0637) |

Note: for all the listed ConvNets, causality images are used as the network inputs; bold values indicate the best results.

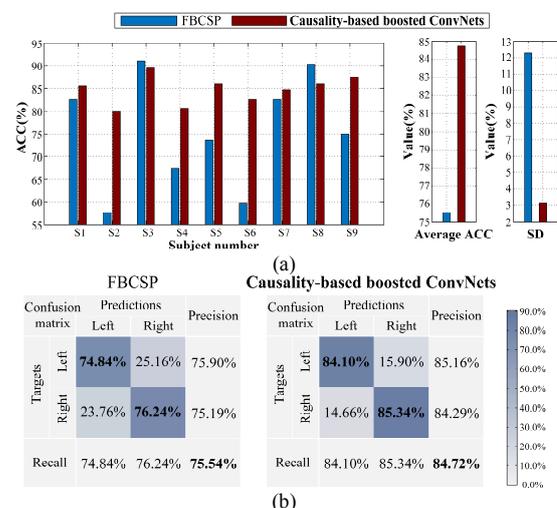

Fig. 3. A comparison of FBCSP and the proposed causality-based boosted ConvNets method: (a) classification results; (b) confusion matrices.

## C. The comparison with FBCSP: ConvNets with causality images reach FBCSP performance

Compared with FBCSP-based decoding, higher classification performance is consistently observed by using ConvNets with causality input images. The classification results and confusion matrices by the proposed method and the FBCSP are given in Fig. 3. For confusion matrices, percentages and colors in the upper-left 2 × 2-square indicate fraction of trials in this cell from all the trials of the corresponding target class; the lower-right value corresponds to overall accuracy, and the bottom row and rightmost column are recall and precision value, respectively.

In detail, the causality-based ConvNets method improves the decoding accuracies of 7 subjects, and the accuracy increases are especially significant for subject 2 (from 57.64% to 79.86%) and subject 6 (form 59.72% to 82.64%) who with poor performance via conventional FBCSP method. Overall, using the proposed ConvNets, the average classification accuracy of FBCSP is enhanced from 75.54% to 84.72%; more importantly, a pretty small inter subject SD value of 3.18% is achieved, which is 74.02% lower than that of the FBCSP. The tremendous decrease of SD suggests that the ConvNets utilizing multi-domain connectivity information of multi-channel EEG signals with special designed spatial and temporal filters, are more robust to individual dependent differences in MI classification than traditional methods, and thus can be applied to various participants. Additionally, the precision and recall of both left- and right-hand tasks are also obviously increased by the proposed networks. The results indicate that the combination of high-resolution causal patterns and ConvNets advantages can indeed improve the performance of MI-EEG decoding.

## D. Classification performance depending on the cropping and boosting strategies

The cropping and boosting strategies are employed in ConvNets to intensify discrimination between causal features of different classes, where dynamical information of single-trial data is maximally exploited. To demonstrate the effectiveness of the cropping and boosting, we compare the causality-based ConvNets using various training strategies in Fig. 4.

It is clear that a great improvement has been made on the test accuracy for almost all the subjects with the adoption of cropped and boosted algorithms. Specifically, with the usage of crops, an accuracy increase of 10.90%, 12.62% and 9.34% is gained on subject 3, 5 and subject 7, respectively; and the average accuracy improves from 77.78% to 80.09%. The employment of boosting strategy can further enhance the classification performance. In detail, except subject 6 with a slight increase of 0.85%, the boosted ConvNets receive the accuracy gains of at least 4.20% for other subjects, besides obtain an average accuracy rise of 5.78%. Note that the inter subject standard deviations by different ConvNets are small and close to each other (3.64% vs. 3.29% vs. 3.18%), which implies that with high-resolution causality input patterns, good robustness can be provided adopting each considered training methods. In addition, the ConvNets using cropping and boosting also arrives at higher recall values while keeping comparative precisions for both two tasks compared to the other networks. The results indicate that the cropping and boosting strategies potentially help to develop more discriminating causal features and thus increase the ConvNets decoding ability.

## E. Analysis of causality images

In the proposed causality-based decoding scheme, the high-resolution time-frequency causality input images for boosted ConvNets are computed from multi-channel EEG data using the multiwavelets-ROFR TF-CGC method. The average of causality images for subject 3 (the best performing subject when the proposed boosted ConvNets is used) are shown in Fig. 5, where the image from original trials of 4 s and that from corresponding cropped samples of 2 s are both illustrated. It is obvious that the significant TF-CGCs from C4 to C3 are more evident than that from C3 to C4 under left-hand conditions (brighter color of C4 compared with C3), while for right-hand, such activations occurred as the opposite electrode C3 exerting relatively stronger causal influences on the same side channel C4 especially within 1-3 s (brighter color of C3 with respect to C4); these estimated dominant causal distributions are consistent with the reported results in MI related researches [41, 42]. It indicates that by applying the proposed TF-CGC method, different connectivity patterns due to interhemispheric brain activations during left- and right-hand MI tasks can be well revealed along the vertical location of the input image.

## IV. DISCUSSIONS

In this paper, a ConvNet framework is proposed for MI decoding in EEG signals, where network input images are based on the time-frequency causality analysis method using multiwavelets and ROFR algorithm. Multiple wavelet basis functions can represent time-varying parameters flexibly, and thus have good generalization properties in describing high-resolution TF-CGC distributions for causal feature extraction. The boosted ConvNets with spatial and temporal convolution blocks are further performed to generate discriminative multi-domain connectivity features from the newly designed causality input images and to train the optimal classifier. Experimental results show that this decoding scheme applying boosted ConvNets together with multiwavelets-ROFR causality method achieves excellent efficiency in MI-EEG classification problem.

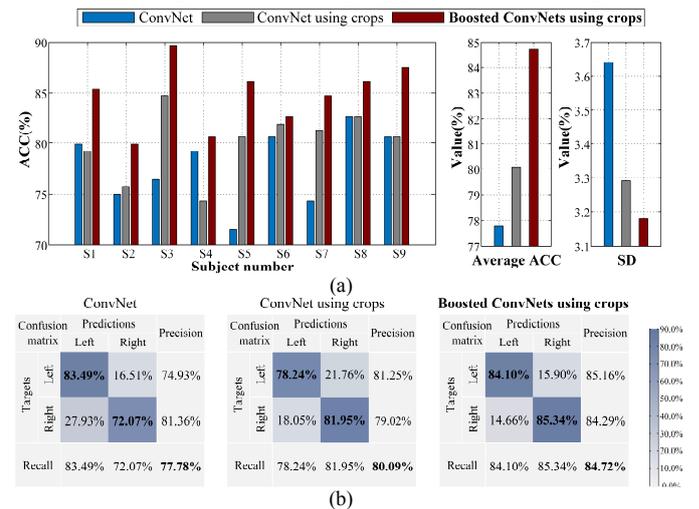

Fig. 4. Classification performance depending on the cropping and boosting strategies: (a) classification results; (b) confusion matrices.

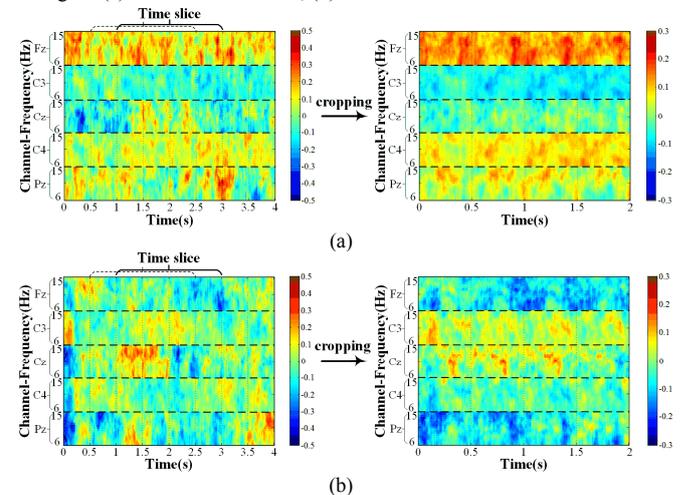

Fig. 5. The average of causality images for subject 3: (a) left-hand MI; (b) right-hand MI.

*A. Efficacy of the multiwavelets and ROFR method*

To evaluate the effectiveness of the multiwavelets and ROFR based causality method, three time-frequency causality approaches are investigated and compared in MI-EEG decoding. Fig. 6 shows the average causality input images calculated by using the adaptive RLS (with forgetting factor 0.98), multiwavelets-OFR and multiwavelets-ROFR methods, respectively. Obviously, the RLS produces poor causality representations and fails to reflect causal influence between C3 and C4 in left-hand MI cases (since no apparent relations appeared in the image), owing to the potential deficiency in slow convergence speed. For measures adopting OFR method, detailed interaction variations cannot be tracked, because the algorithm could still attempt to fit noisy data although meanwhile try to follow the parsimonious principle, thus makes the time-varying models over-fitting and leads to spurious causal values. By contrast, the proposed multiwavelets-ROFR method can avoid the slow convergence limitation and over-fitting when modelling time-varying signals, and thereby higher resolution and more accurate time-frequency causal distributions can be achieved.

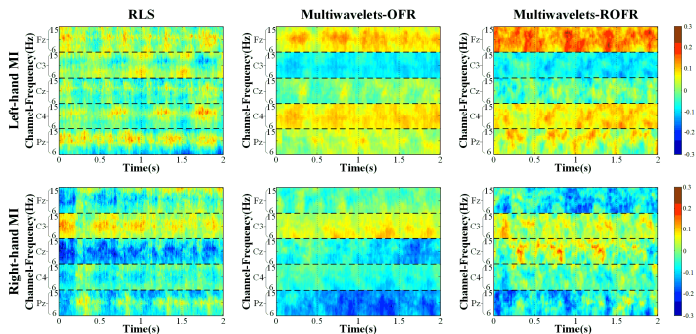

Fig. 6. The average of causality input images by different TF-CGC methods.

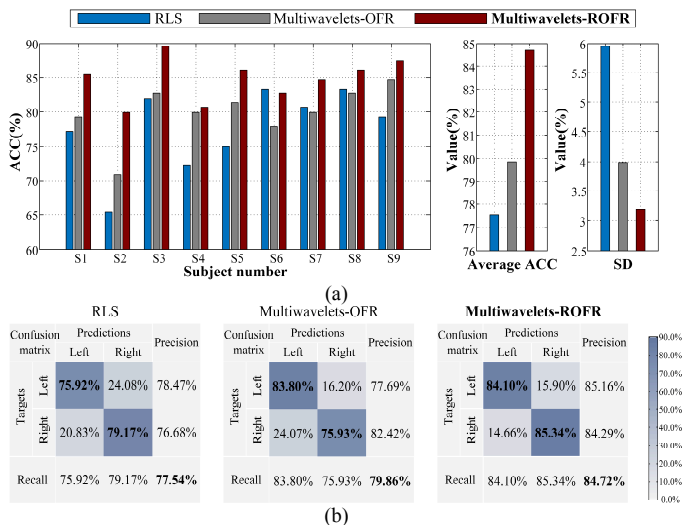

Fig. 7. Classification performance effected by multiwavelets-ROFR based TF-CGC method: (a) classification results; (b) confusion matrices.

A comparison of the classification performance of ConvNets with different causality methods is given in Fig. 7. Specifically, using the multiwavelets-OFR approach, classification accuracies of 5 subjects are increased at least 2.71% compared to the RLS; the average accuracy is improved from 77.54% to 79.86%, and the inter subject standard deviation is dropped from 5.95% to 3.99%. A good rise on the accuracy is further obtained for every subject by the proposed multiwavelets-ROFR method. In detail, the ROFR ConvNets receives an average accuracy increase of 6.09% and a standard deviation decrease of 20.30% than the OFR. Moreover, the proposed method also reaches the highest precision and recall for either classification tasks among these causality methods. The improvement of classification results indicates that the multiwavelets-ROFR based causality analysis can indeed detect high-resolution TF-CGCs from nonstationary EEG signals and increase the capacity of ConvNets in MI decoding.

The superiority of the proposed multiwavelets-ROFR approach is partly attributed to the ROFR algorithm which refines redundant model terms and constructs a sparse model with good generalization ability for nonstationary time series. Thus high-precision TF-CGC images can be calculated by not only the global frequency behavior but also the local causal variations of EEG signals over time, and a higher decoding performance can be finally obtained by the proposed method.

*B. Efficacy of the newly designed causality input images for deep learning models*

It is noteworthy that the proposed ConvNets with causality input images can significantly decrease inter subject standard deviations compared with the classic FBCSP algorithm. This shows that by employing subject specific training on high-resolution TF-CGC images, ConvNets can overcome the large difference between classification performance of participants which is important for MI-BCI applications. It is the novel multi-domain (time, frequency and space) causality inputs, that reveal dynamical patterns of *alpha* band brain connectivity activities underlying motor perceptual decisions, that makes the ConvNet model more robust than conventional methods. Deep learning models for EEG analysis generally adopt the following two input styles: 1) the EEG signals of all available channels, and 2) the transformed EEG signals (such as a TF decomposition) of all available or a subset of channels. The proposed network uses the second input form, that is, time-frequency causal representations calculated from multi-channel EEG data. To illustrate the efficiency of causality input images for ConvNet in EEG decoding, the proposed method and the deep and shallow ConvNets [27], which fall in the first input style, are compared in Fig. 8 and Table 2.

Despite that the classification accuracy varies for each individual and that the confusion matrix of the proposed method is similar to that of the shallow ConvNet, an average accuracy increase of 18.32% and 2.33% is gained with respect to the deep and shallow ConvNet, respectively, by using the ConvNets with causality images. More importantly, our causality-based method shows a more stable decoding performance than others. Specifically, the difference of accuracy between the best subject (subject 3, ACC = 89.58%) and the worst one (subject 2, ACC = 79.86%) is 9.72% when the proposed method is used, whereas this value is 46.53% for deep ConvNet and 40.98% for the shallow. The standard deviation in this study is 3.18%, which gets 83.40% decrease than deep ConvNet, and reduces 80.17% to the shallow method.

The results indicate that compared with the ConvNets directly using EEG signals as model inputs, the newly proposed architecture taking advantage of multi-domain connectivity

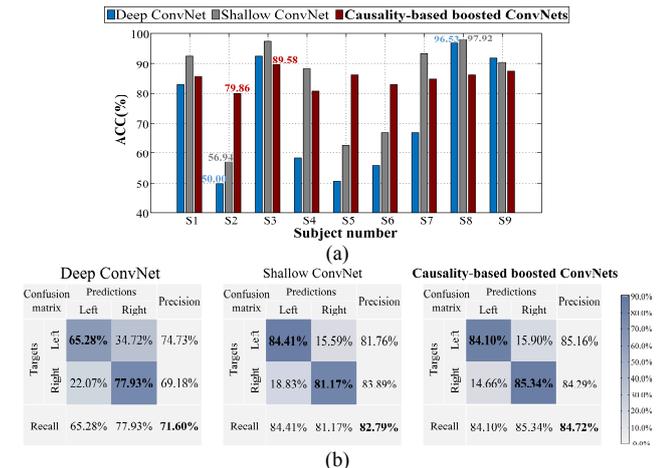

Fig. 8. Classification performance effected by causality input images: (a) classification results; (b) confusion matrices.

TABLE II
DECODING PERFORMANCES OF DIFFERENT CONVNETS

| Method | Frequency range [Hz] | The number of used channels | Average ACC (%) | SD (%) |
|---|---|---|---|---|
| Deep ConvNet | 4-38 | 22 | 71.60 | 19.16 |
| Shallow ConvNet | 4-38 | 22 | 82.79 | 16.04 |
| **Boosted ConvNets with causality images** | **6-15** | **5** | **84.72** | **3.18** |

Note: bold values indicate the proposed method.

features convolved from causality inputs can yield more uniformly distributed accuracies with no extremes, and thus has a great potential to solve the problem of subject dependent disturbances in MI-BCI systems. Furthermore, note that the proposed ConvNets outperforms other methods by analyzing more narrow frequency band and much less EEG channels, implying abundant and effective information contained in the causal features for the discrimination of motor perceptions.

*C. Efficacy of cropped training*

In the proposed method, the cropped training strategy for data augmentation is essential for achieving the competitive classification accuracy in MI-EEG decoding. The accuracy improvement due to the use of crops indicates that a large number of training samples is necessary for the ConvNets to learn to extract discriminating connectivity features. This makes sense as the shifted neighboring windows can contain the same, but moving, oscillatory causality information. These shifts can take full advantage of *alpha* band dynamic causal patterns during the whole 4 s reaction period, and prevent the network from overfitting on phase information within the trial. In addition, the trained network can be more adaptable for testing since more variety of cropped inputs are generated for training.

*D. Efficacy of Adaboost algorithm*

The decoding performance has been further improved by re-weighting the cropped training samples employing Adaboost algorithm. This suggests that the initial augmented causal features may be redundant because of the overlapped information within crops; while this redundancy issue can be well solved by adaptive adjustment of the training data distribution which further selects more discriminative features during the iterative training. Moreover, our experiment results show that, for such a boosted network, using a small number of iterations (when $\chi$ reaches around 20) can results in an obvious improvement of accuracy, indicating that the temporal patterns of causal features preliminarily extracted by the constructed ConvNets are essentially discriminatory; this reflects the effectiveness of the proposed architecture in feature extraction.

*E. Effectiveness of the causality-based boosted ConvNets*

Apart from the discriminative multi-domain causality input images, another key to the classification performance improvement lies in the application of the recent developments in deep learning. Specifically, the use of batch normalization, dropout and ELUs in the proposed ConvNets leads to the accuracy increasing. To explain the superiority of the proposed causality-based boosted ConvNets, a comparison between the proposed decoding scheme and the state-of-the-art methods published for MI-EEG classification is shown in Table 3. All the results are tested on the same EEG dataset for the same binary classification task, thus the comparison is feasible.

It is clear from Table 3 that the proposed method has a considerably robust performance over the nine subjects and the overall accuracy is superior in the comparison. The methods introduced in [51-53] were based on feature extraction methods including the modified CSP, covariate shift-detection, and current source density (CSD), with traditional classification algorithms such as SVM and Bayesian linear discriminant analysis (BLDA). These machine-learning approaches have reached an average accuracy exceeding 80%, but were still lower than our ConvNet method; and a much larger inter subject standard deviations ranged from 10.00% to 14.00% was generated. Additionally, for the deep learning techniques recently suggested by Schirrmeister *et al.* [27], the shallow ConvNet arrived at a relatively high average accuracy of 82.79%, nevertheless the standard deviations of these networks were even slightly larger than the machine-learning schemes reported in [51-53], which indicates a oscillatory decoding performance of the associated deep learning models. This further emphasizes the crucial role of connectivity patterns for the proposed deep network to produce the steady classification results over different subjects. In general, the results suggest that the proposed causality-based boosted ConvNets can be regarded as an effective tool for MI-EEG decoding.

TABLE III
A COMPARISON OF VARIOUS METHODS FOR MI-BCI CLASSIFICATION

| | Methods | | | | | |
|---|---|---|---|---|---|---|
| | Bayesian spatio-spectral filter optimization and SVM (2013) [51] | Adaptive learning with covariate shift-detection (2016) [52] | CSD (Current Source Density) and SVM (2017) [53] | Deep ConvNet (2017) [27] | Shallow ConvNet (2017) [27] | **This work** |
| S1 | **93.75** | 85.71 | 93.06 | 82.64 | 92.36 | 85.42 |
| S2 | 63.19 | 75.71 | 68.06 | 50.00 | 56.94 | **79.86** |
| S3 | **98.61** | 92.86 | 93.06 | 92.36 | 97.22 | 89.58 |
| S4 | 70.14 | 77.86 | 77.08 | 58.33 | **88.19** | 80.56 |
| S5 | 76.39 | 61.43 | 72.22 | 50.69 | 62.50 | **86.11** |
| S6 | 74.31 | 71.43 | 65.97 | 55.56 | 66.67 | **82.64** |
| S7 | 86.11 | 84.29 | 78.47 | 66.67 | **93.05** | 84.72 |
| S8 | 96.53 | 93.57 | 97.22 | 96.53 | **97.92** | 86.11 |
| S9 | **95.14** | 80.00 | 90.28 | 91.67 | 90.28 | 87.50 |
| Mean (SD) | 83.80 (13.09) | 80.32 (10.25) | 81.71 (11.87) | 71.60 (19.16) | 82.79 (16.04) | **84.72 (3.18)** |

Note: bold values indicate the best results.

*F. Limitations and potentials for improvement*

There are mainly two limitations in the current work. First, the studied frequency range and channel numbers may affect the accuracy of the proposed ConvNets method. Specifically, 6-15 Hz (*alpha* band) oscillatory activity of 5 channels are used in this work, which can be extended to a wider band and more available channels in future. With more connectivity information in frequency and location domain being extracted, the decoding performance of the causality-based ConvNets can be promoted. The second limitation is the size of the dataset. Although the cropping strategy is applied in this work to augment data, the number of samples is still quite small compared with the commonly used datasets for deep learning. Moreover, the potential of deeper ConvNets in EEG decoding might not be fully explored because of the limited amount of data. This dataset-size problem could be alleviated by transfer learning approaches across participants and other datasets in future, so that the efficiency of the proposed scheme can be further improved.

## V. Conclusion

In this paper, a novel decoding scheme is proposed for MI-EEG signals using boosted ConvNets with the multiwavelet-based TF-CGC images. The dynamic TF-CGC approach implemented by multiwavelets-ROFR can produce high-resolution time-varying spectral causality distributions from nonstationary EEG signals. The boosted ConvNets, which utilize convolutional models and deep learning methods with cropping and boosting strategies, can extract discriminative connectivity features from the newly designed multi-domain causality input images, and accomplish the classification of MI tasks. The effectiveness of the proposed method is evaluated and compared with classic FBCSP and the networks employing various training approaches. Experimental results show that, with the usage of crops and the boost algorithm, our ConvNets achieves the best decoding performance in the binary classification case between left- and right-hand MI-EEGs, and the average accuracy and kappa are 84.72% and 0.6944 with inter subject standard deviation of 3.18% and 0.0637, respectively. The efficacy of the multiwavelets-ROFR causal approach is also validated with the proposed decoding architecture, and significant improvement of classification performance is attained as to RLS and multiwavelets-OFR methods. Furthermore, our overall classification results on the public MI-EEG dataset are better than the compared state-of-the-art methods, especially in terms of the inter subject standard deviation metric. One obvious advantage of the proposed scheme is that multi-domain connectivity information characterized by deep learning techniques is used to represent different MI tasks; in this way it makes good use of the variations in causality images to generate discriminative features, and further obtains reliable and stable decoding performance for various participants. Another advantage is that multiwavelets and ROFR are integrated and applied to better approximate TVARX models to enhance generalization capability, which can provide high-resolution and physiologically accordant TF-CGC patterns from EEG data. It is expected that the proposed ConvNets decoding scheme can improve the interpretability of deep learning models according to the underlying connectivity activities within electrophysiological networks, and further enhance the robustness and applicability of MI-BCI systems.

## Appendix

*A. Algorithm 1: Pseudocode for sparse model structure determination using ROFR algorithm*

**Algorithm 1:** Pseudocode for sparse model structure determination using ROFR algorithm

**Input:**
output vector $X = [x(1), x(2), \cdots, x(N)]^T$; candidate terms $W = \{\xi_1, \xi_2, \cdots, \xi_M\}$; regularization parameter $\rho$; adjustable parameter for PESR criterion $\mu$; predefined threshold $\epsilon = 10^{-\ell}$ where $\ell > 10$.

**Initialize:**
Set $I_1 \leftarrow \{1, 2, \cdots, M\}$; $r_0 \leftarrow X$.

*Step 1:*
for $m = 1, 2, \cdots, M$ do
$$h_m^{(1)} = \xi_m,\ RERR_m = \frac{(\xi_m^T r_0)^2}{r_0^T r_0 (\xi_m^T \xi_m + \rho)}$$
end for
$L_1 \leftarrow \arg\max_{m \in I_1}\{RERR_m\}$, $h_1 = h_{L_1}^{(1)}$, $r_1 = r_0 - \left(\frac{r_0^T h_1}{h_1^T h_1}\right) h_1$

*Step $\varsigma$ ($\varsigma \geq 2$):*
for $\varsigma = 2, 3, \cdots, M$ do
  $I_\varsigma \leftarrow I_{\varsigma-1} \setminus \{L_{\varsigma-1}\}$
  for all $m \in I_\varsigma$ do
  $$h_m^{(\varsigma)} = \xi_m - \sum_{v=1}^{\varsigma-1}\left(\frac{\xi_m^T h_v}{h_v^T h_v}\right) h_v,\ RERR_m = \frac{\left(h_m^{(\varsigma)} r_{\varsigma-1}\right)^2}{r_{\varsigma-1}^T r_{\varsigma-1}\left(\left(h_m^{(\varsigma)}\right)^T h_m^{(\varsigma)} + \rho\right)}$$
  end for
  $\mathcal{K}_\varsigma \leftarrow \left\{\arg_{m \in I_\varsigma}\left(h_m^{(\varsigma)}\right)^T h_m^{(\varsigma)} < \epsilon\right\}$, $I_\varsigma \leftarrow I_\varsigma \setminus \mathcal{K}_\varsigma$
  $L_\varsigma \leftarrow \arg\max_{m \in I_\varsigma}\{RERR_m\}$, $h_\varsigma = h_{L_\varsigma}^{(\varsigma)}$, $r_\varsigma = r_{\varsigma-1} - \left(\frac{r_{\varsigma-1}^T h_\varsigma}{h_\varsigma^T h_\varsigma}\right) h_\varsigma$
  $PESR(\varsigma) = \frac{1}{(1-\mu\varsigma/N)^2}\left(1 - \sum_{m=1}^{\varsigma} RERR_{L_m}\right)$
end for
$q = \arg\min_{\varsigma \in I_1}\{PESR(\varsigma)\}$

**Output:**
selected model terms $\Phi = \left[\xi_{L_1}, \xi_{L_2}, \cdots, \xi_{L_q}\right]$.

*B. Algorithm 2: Pseudocode for boosted ConvNets based on AdaBoost algorithm*

**Algorithm 2:** Pseudocode for boosted ConvNets based on AdaBoost algorithm

**Input:**
training dataset $\{(X_1, y_1), \cdots, (X_{Y/2}, y_{Y/2})\}$; the number of iterations $\chi$.

**Initialize:**
Set $N_T \leftarrow 0.8 \times \frac{Y}{2}$; $N_V \leftarrow 0.2 \times \frac{Y}{2}$; $\omega_1(X_\gamma) \leftarrow \frac{1}{N_T}$, $\gamma = 1, 2, \cdots, N_T$; $ACC_{Val} \leftarrow 0$.

for $j = 1, 2, \cdots, \chi$ do
  Training deep ConvNet model $\Lambda_j$ with $\{(\omega_j(X_\gamma)X_\gamma, y_\gamma), \gamma = 1, 2, \cdots, N_T\}$
  $\Xi_j = \sum_{\gamma=1}^{N_T} \omega_j(X_\gamma) I(\Lambda_j(X_\gamma) \neq y_\gamma)$, $\varpi_j = \frac{1}{2}\ln\frac{1-\Xi_j}{\Xi_j}$
  $\Omega_j = \sum_{\gamma=1}^{N_T} \omega_j(X_\gamma) \exp\left(-\varpi_j y_\gamma \Lambda_j(X_\gamma)\right)$
  for $\gamma = 1, 2, \cdots, N_T$ do
    $\omega_{j+1}(X_\gamma) = \frac{\omega_j(X_\gamma)}{\Omega_j} \exp\left(-\varpi_j y_\gamma \Lambda_j(X_\gamma)\right)$
  end for
  $\Lambda = \sum_{i=1}^{j} \varpi_i \Lambda_i$, $acc \leftarrow \frac{1}{N_V}\sum_{\gamma=N_T+1}^{Y/2} I(\Lambda(X_\gamma) = y_\gamma)$
  if $acc > ACC_{Val}$
    $\tilde{\Lambda} \leftarrow \Lambda$, $ACC_{Val} = acc$
  end if
end for

**Output:**
final boosted ConvNets model $\tilde{\Lambda}$.

*C. Fig. 2: Illustration for TF-CGC calculation.*

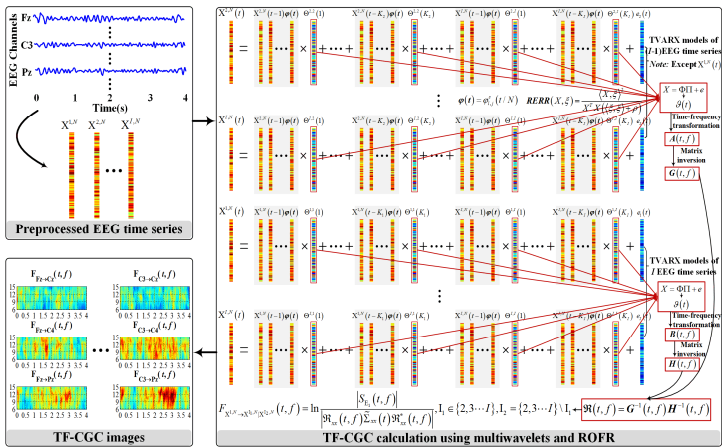

Fig. 2. Illustration for TF-CGC calculation.